\documentclass[letterpaper, 10 pt, conference]{ieeeconf}  

\IEEEoverridecommandlockouts                              
\usepackage{verbatim}
\usepackage{mathtools}
\usepackage{amssymb}
\usepackage{enumerate}
\usepackage{tikz}
\usepackage{tikzscale}
\usepackage[hidelinks]{hyperref}
\usepackage[multiple]{footmisc}
\usepackage{graphicx}
\usepackage{array}
\usepackage{gensymb}
\usepackage{cite}
\usetikzlibrary{positioning}
\usetikzlibrary{shapes.geometric, arrows}
\let\savedegree\degree
\let\degree\relax
\usepackage{mathabx}
\usepackage{booktabs}
\let\degree\savedegree

\newcommand{\updrive}{\textit{\mbox{UP-Drive} }}
\newcommand{\updrivens}{\textit{\mbox{UP-Drive}}}

\newcommand{\bo}{B}
\newcommand{\ma}{M}

\newcommand{\poseab}[2]{T^{#1}_{#2}}
\newcommand{\poset}[1]{\poseab{#1}{\ma \bo}}

\newcommand{\poseguess}[1]{\hat{T}^{#1}_{\ma \bo}}

\newcommand{\recallmt}{\mathbf{r}_{mt}}

\newcommand{\planarmedianerror}{\mathbf{\bar{p}^{e}_{xy}}}
\newcommand{\planarerror}{\mathbf{p^{e}_{xy}}}
\newcommand{\latmedianerror}{\mathbf{\bar{p}^{e}_{y}}}
\newcommand{\laterror}{\mathbf{p^{e}_{y}}}
\newcommand{\orientmedianerror}{\mathbf{\bar{\theta}^{e}_{xyz}}}
\newcommand{\orienterror}{\mathbf{\theta^{e}_{xyz}}}

\newcommand{\timestep}{t}

\newcommand{\villoc}{MOZARD }
\newcommand{\visloc}{VIZARD }
\newcommand{\vislocns}{VIZARD}
\newcommand{\villocns}{MOZARD}
\newcommand{\sevensensesymbol}{2}
\newcommand{\aslsymbol}{1}
\newcommand{\ethsymbol}{3}

\tikzstyle{startstop} = [rectangle, rounded corners, minimum width=3cm, minimum height=1cm,text centered, draw=black, fill=red!30]
\tikzstyle{io} = [trapezium, trapezium left angle=70, trapezium right angle=110, minimum width=0cm, minimum height=0cm, text centered, draw=black, fill=blue!30]
\tikzstyle{process} = [rectangle, minimum width=0cm, minimum height=0cm, text centered, text width=3cm, draw=black, fill=orange!30]
\tikzstyle{decision} = [diamond, minimum width=0cm, minimum height=0cm, text centered, draw=black, fill=green!30]
\tikzstyle{arrow} = [thick,->,>=stealth]

\title{\LARGE \bf
\villocns: Multi-Modal Localization for Autonomous Vehicles in Urban Outdoor Environments
}

\author{Lukas Schaupp$^{\aslsymbol}$, Patrick Pfreundschuh$^{\ethsymbol}$, Mathias B\"{u}rki$^{\aslsymbol,\sevensensesymbol}$, Cesar Cadena$^{\aslsymbol}$,\\ Roland Siegwart$^{\aslsymbol}$, and Juan Nieto$^{\aslsymbol}$
\\ \small$^{\aslsymbol}$Autonomous Systems Lab, ETH Z\"{u}rich, {\tt\footnotesize \{firstname.lastname\}@mavt.ethz.ch}
\\
$^{\sevensensesymbol}$Sevensense Robotics AG, {\tt\footnotesize \{firstname.lastname\}@sevensense.ch}
$^{\ethsymbol}$ETH Zurich, {\tt\footnotesize \{firstname.lastname\}@student.ethz.ch}
}

\begin{document}

\maketitle
\thispagestyle{empty}
\pagestyle{empty}

\begin{abstract}
%

Visually poor scenarios are one of the main sources of failure in visual localization systems in outdoor environments.
To address this challenge, we present \villocns, a multi-modal localization system for urban outdoor environments using vision and LiDAR.
By extending our preexisting key-point based visual multi-session local localization approach with the use of semantic data, an improved localization recall can be achieved across vastly different appearance conditions. In particular we focus on the use of curbstone information because of their broad distribution and reliability within urban environments.
%
%
We present thorough experimental evaluations on several driving kilometers in challenging urban outdoor environments, analyze the recall and accuracy of our localization system and demonstrate in a case study possible failure cases of each subsystem. 
We demonstrate that \villoc is able to bridge scenarios where our previous work \visloc fails, hence yielding an increased recall performance, while a similar localization accuracy of 0.2$m$ is achieved.
\end{abstract}

\section{Introduction}
Due to increasing traffic in urban environments and changing customer demands, self-driving vehicles are one of the most discussed and promising technologies in the car and robotics industry. Still no system was presented yet, that allows to localize robustly under all light, weather and environmental conditions. However, precise localization is a vital feature for each autonomous driving task, since a wrong pose estimate may lead to accidents. Especially in urban environments, safety margins on the position of the car are small due to crowded traffic and other traffic participants (e.g. pedestrians, cyclists).
Because of multi-path effects or satellite blockage, GPS sensors cannot be used reliably under those urban conditions. Thus, other sensors have to be used for localization. For this purpose, mainly LiDARs and cameras have been used in the last years. 
Appearance changes in urban environments challenge visual localization approaches. However, such driving scenarios contain persistent structures even under those appearance changes. Curbstones are one such feature.
%
%
Curbstones are used to protect pedestrians from cars and to separate the sidewalk from the street. As they delimit the street, they also offer information of the area where the car is allowed to be placed in. Detection of their position relative to the car can thus allow to localize inside the lane.
In contrast to other geometrical shapes such as poles and road markings, curbstone measurements are found more frequently in urban environments and yield a reliable, continuous lateral constraint for pose refinement.
Due to their shape and their contrasting color with respect to the pavement in many cases, they can be detected both in camera images as well as in LiDAR pointclouds. 
%
\begin{figure}
\includegraphics[width=0.49\textwidth]{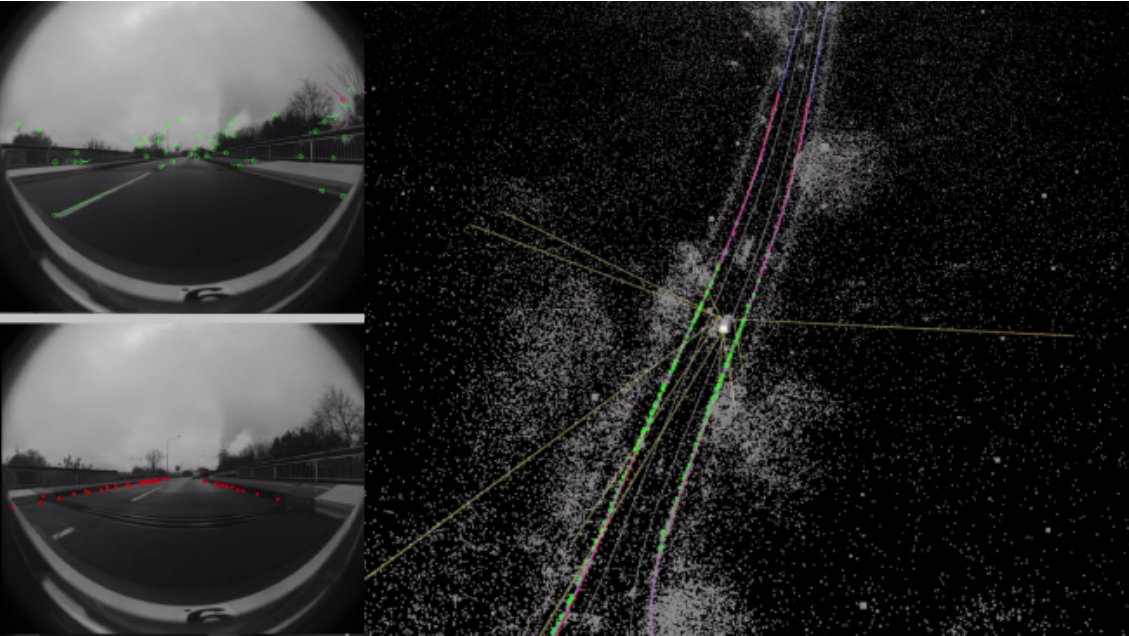}
\caption{\label{fig:teaser}
We aim at accurately localizing the \updrive vehicle in a map of features extracted from vision and LiDAR data depicted on the right side. Our proposed algorithm can be separated into two distinct steps.
We extract keypoint-based features from camera and additional 3D geometrical curbstone information from a semantic vision-LiDAR pipeline. The features extracted from images of the surround-view camera system (top-left corner) are matched against $3D$ landmarks in the map while our raw curbstone measurements (bottom-left corner) are matched to their corresponding landmarks.
Inlier matches, centered on the estimated $6DoF$ pose of the vehicle in the map, are illustrated as dark yellow lines on the right side. 
Purple indicates our pre-generated curbstone map data represented by splines. During runtime we downsample the nearest splines spatially to match with the current raw curbstone measurements indicated by the red and green color.
}
\vspace{-4mm}
\end{figure}
Therefore, our pipeline named \villoc extends our previous visual localization system - \vislocns \cite{brki2019vizard} with the use of additional geometrical features from LiDAR data, for the self-driving cars used in the \updrive project\footnote{The \textbf{UP-Drive} project is a research endeavor funded by the European Commission, aiming at advancing research and development towards fully autonomous cars in urban environment. See \url{www.up-drive.eu}.}.
In a thorough evaluation of our proposed localization system using our long-term outdoor dataset collection, we investigate key performance metrics such as localization accuracy and recall and demonstrate in a case study possible failure scenarios.
%
%
%
%
%

We see the following aspects as the main contributions of this paper:
\begin{itemize}
\item A semantic extension of our key-point-based localization pipeline based upon the extraction of curbstone information is presented, that allows to bridge sparse key-point based feature scenarios in visual localization.
\item In a thorough evaluation on the long-term dataset collection \updrivens, we demonstrate a reliable localization performance across different appearance conditions in urban outdoor environments. We compare our results to our vision-based localization pipeline and demonstrate significant performance increases.
\item A computational performance analysis showing that our proposed algorithm exhibits real-time capabilities and better scalability. 
%
%
\end{itemize}

\section{related work}
Since our localization system is a multi-modal semantic extension of our previous work we will concentrate the related work on frameworks that exploit semantic features using either one modality or fusing multiple modalities in different ways. Therefore these works can be subdivided into each of their specific sensor setup. For general related work on our prior visual key-point based system we refer to our previous work \cite{brki2019vizard}.

\subsubsection*{Vision-only}
\label{sec:related_work:subsec:local_localization}
A recent example of using semantic features for the purpose of localization and mapping is Lu et al. \cite{Lu2017} using a monocular camera for road-mark detection whereas other studies use traffic signs \cite{DOrazio2018}, line segments from multi-view cameras \cite{Hara2015} or poles \cite{Weng2018} \cite{Spangenberg2016} for feature matching. On the detection of curbstones, traditional image based curbstone detection mostly uses the vanishing point and color distribution to detect the corresponding pixels \cite{traditionalimage}, \cite{lanelocalization}. Recent work such from Enzweiler et al. \cite{Enzweiler2013} and Panev et al. \cite{Panev2019} also demonstrate a learning based approach to detect curbs in images. In contrast to the vision based approaches, we concentrate on the multi-modal aspect as provided by Goga et al. \cite{Goga2018}.
\subsubsection*{LiDAR-only}
\label{sec:related_work:subsec:local_localization}
LiDAR based methods use assumptions about the shape of the semantic features like curbs, poles and planes by evaluating the difference in elevation \cite{Liu13anew, miraliakbari}, slope \cite{Liu13anew} or curvature \cite{Lopez2015, Lopez2014}. Authors such as Schaefer et al. \cite{Scheafer2019} detect and extract 3D poles from the scenery which are then being used for map tracking. In regards to the usage of curbstones, most applications use geographical map data \cite{Bonnifait2008,Qin} or road networks \cite{stueckler} as a reference to localize with detected curbstones. Unlike these works, our approach does not rely on external pre-generated data for curbstone map construction.

\subsubsection*{Vision and LiDAR}
\label{sec:related_work:subsec:local_localization}
%
Recent work such from Kampker et al. \cite{Kampker2019} use a camera to extract pole like landmarks and a LIDAR for cylinder shapes for the task of self-localization. Kummerle et al. \cite{Kummerle2019} demonstrate that basic geometric primitives can be extracted using vision and LiDAR to obtain road markings, poles and facades which can then be used for localization and mapping for the purpose of self-localization on various weather conditions. While these approaches use both modalities for mapping and localization separately, there has been recent research into cross-modality. Xiao et al. \cite{Xiao2018} uses a LiDAR to build an HD map and extract 3D semantic features from the map. Then a monocular camera is used with a deep learning based approach to match these semantic features with the ones from the camera. 
%
In contrast to the graph-based SLAM formulation used in our approach, the mentioned approaches are filter-based, with Extended Kalman Filters \cite{Lee2014} or Monte Carlo Localization \cite{Qin, Kampker2019}, and they are evaluated at low speed and/or on short maps of a few hundred meters \cite{Xiao2018}. In addition they do not use raw curbstone measurements as a feature for localization and mapping \cite{Kummerle2019, Xiao2018, Scheafer2019,Bonnifait2008,Qin}. Our work is evaluated on a long-term map with a length of over 5$km$ using urban driving speed of around 50$km/h$. 
\section{Methodology}
\label{sec:methodology}
\begin{figure}
\includegraphics[width=0.49\textwidth]{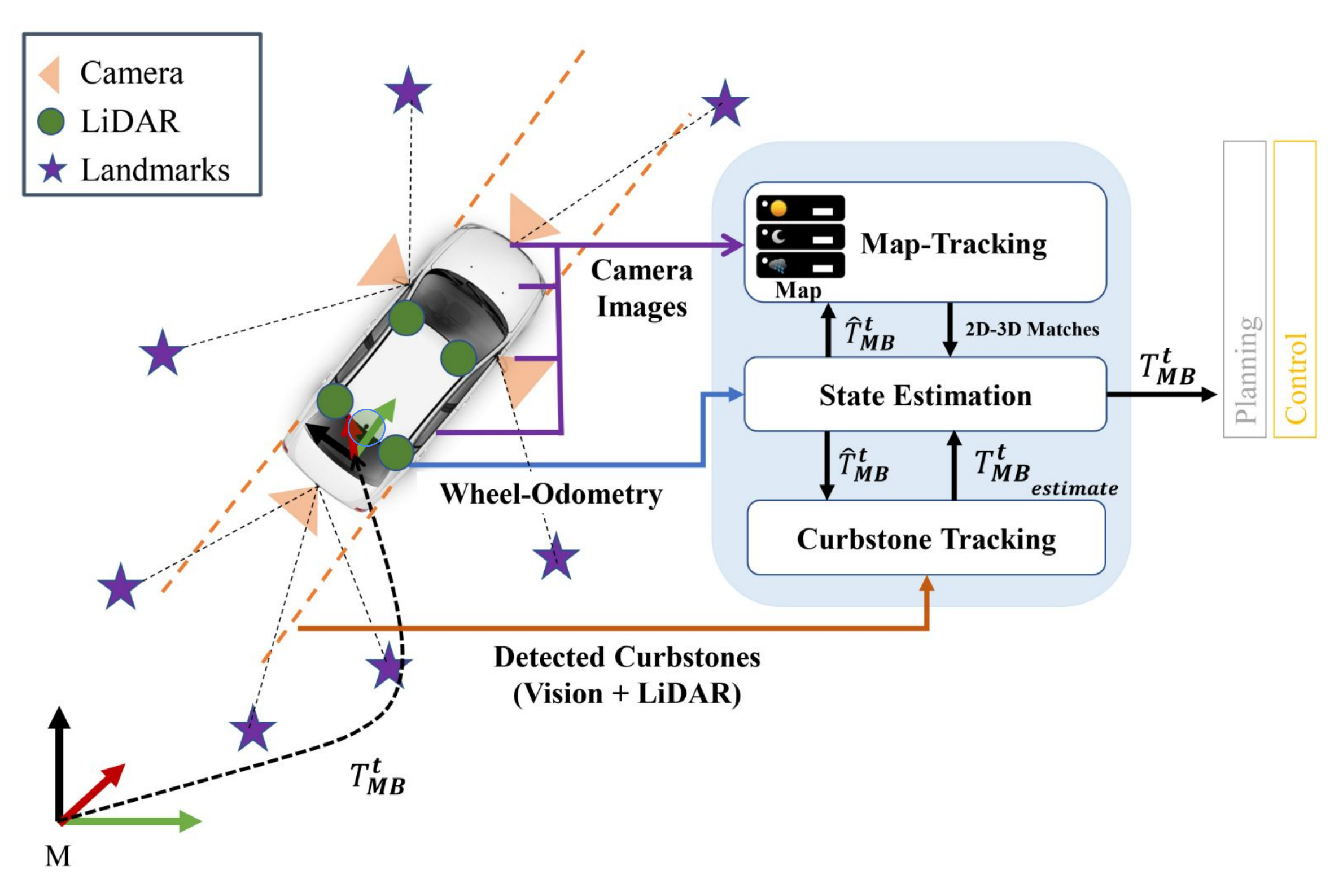}
\caption{\label{fig:schema}
The key-point based map-tracking module extracts 2D features from current camera images, and matches them with 3D map landmarks locally in image space using a pose prior $\poseguess{\timestep}$ while our semantic map-tracking module matches point-cloud based curbstone measurements between our map and immediate input.
The state estimation module fuses the visual \mbox{2D-3D} and geometrical \mbox{3D-3D} matches with the current wheel-odometry measurement to obtain a current vehicle pose estimate $\poset{\timestep}$.
%
}
\end{figure}
A schematic overview of \villoc can be found in Figure~\ref{fig:schema}. Since our work extends the \visloc framework, we refer to the general methodology from Buerki et al. \cite{brki2019vizard}. We assume that our visual localization pipeline already created a map by tracking and triangulating local 2D features extracted along a trajectory.

\subsection{Curbstone Detection}
\label{sec:methodology:subses:swe}
For our curbstone detection we employ the work from Goga et al. \cite{Goga2018}. Goga et al., fuse a vision-based segmentation CNN with LiDAR data. In a post-processing step they extract, refine and filter semantic curb ROIs to obtain new curb measurements. In the following we use their curbstone detection as input into our pipeline.

\subsection{Map Extension}
\label{sec:methodology:subses:swe}
The curbstones are added to a map that was built using the \visloc pipeline \cite{brki2019vizard}. This map is called the base map in the following. Curbstone points detected in a specific LiDAR pointcloud frame will be called curbstone observation. The detection pipeline finds curbstone pointclouds in the vehicle frame $\mathcal{F}_B$. We find the closest vertex in time within the base map and allocate the curbstone pointcloud. This is performed for all curbstone detections along the trajectory. From the base map, the respective transformations from the map coordinate frame $\mathcal{F}_M$ to the body frame at time $t$ can be looked up. Using $T_{MB}^t$ at each vertex in the base map that contains a curbstone observation, a curbstone map in $\mathcal{F}_M$ can then be created. 

\subsection{Curbstone Map-Tracking}
\label{sec:methodology:subsec:map_tracking}
\begin{figure}
\includegraphics[width=0.49\textwidth]{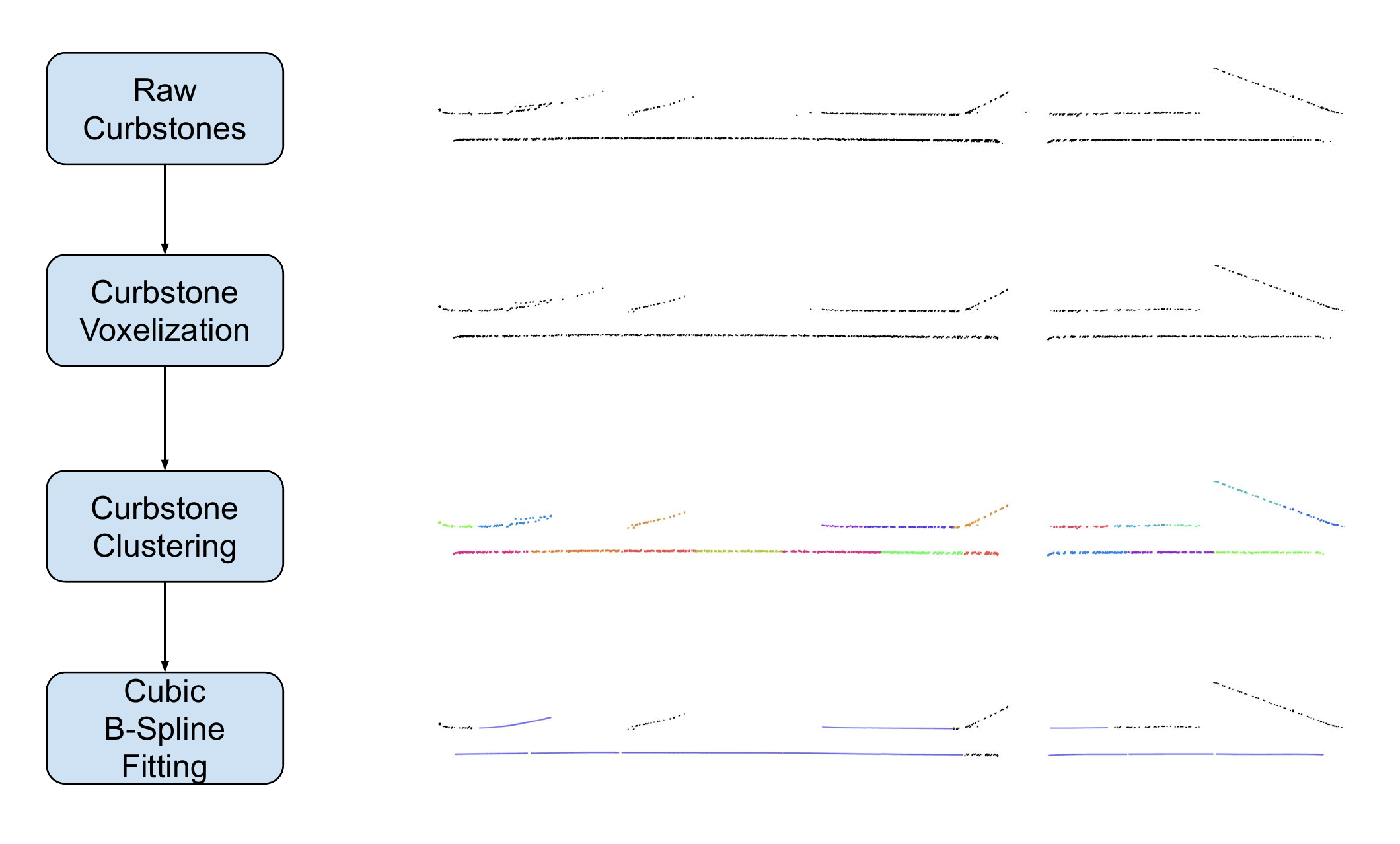}
\caption{\label{fig:schema_curbstone_param} For our curbstone parameterization pipeline we first obtain raw curbstone measurements. After a voxelization processing step, we cluster the curbstones into segments. Finally, a cubic b-spline is fitted to each segment. In case the fitting algorithm fails, we store the raw points.}
\end{figure}
The curbstone tracking module is the core component of the curbstone localization pipeline. It performs an alignment between the map curbstones and the input curbstones to estimate the vehicle pose. A sanity check is performed to detect wrong alignments. If it is fulfilled, a pose constraint is added to the graph. The integration into the \visloc system is shown in Figure \ref{fig:schema}. The single steps are explained in detail in the following section.

\subsubsection{Reference Curbstone Retrieval}
\label{sec:methodology:subsec:mapping}
To retrieve the map curbstones, a prior estimate $\hat{T}^t_{MB}$ is used. In a fixed radius $r_{lookup}$ around the estimated position $\hat{P}^t$, we then search for the closest vertex in Euclidean distance in the base map that contains curbstones. Furthermore, a criterion on the maximum yaw angle between the prior pose estimate and the base map vertex pose is used to prevent wrong associations.
\subsubsection{Pointcloud Registration}
\label{sec:methodology:subsec:mapping}
Given the map and input curbstones a pointcloud registration is performed. As the registration algorithm NDT (Normal Distribution Transform) \cite{Magnusson2009} is used from the implementation of the Point Cloud Library \cite{pcl}. Additionally, outlier points are removed using a fixed ratio since some artifacts might only be included in one of both pointclouds, due to occlusion or unsimilar detection. The pointcloud registration estimates a transformation $T_{align}$ that aligns the input cloud to the map cloud.

\subsubsection{Sanity Check}
\label{sec:methodology:subsec:mapping}
In some regions, the input or map pointcloud can consist of very few points. Matching in those scenarios can be ambiguous and lead to wrong associations. Thus, matching is only performed if both pointclouds exceed a minimum amount of points. Since urban street scenarios change frequently, e.g due to constructions or parked cars, the input and map pointcloud can diverge heavily. In those cases, pointcloud registration might fail, ending up in wrong alignments and thereby wrong pose estimates. Therefore, a sanity check has to be performed, to detect wrong pose estimates. To do so, a matching score can be calculated, that can be used as an indicator if the alignment was successful. 
Magnusson et al. \cite{Magnusson2009} proposed a matching score for NDT. It corresponds to the likelihood that the aligned input points lie on the reference scan. A more detailed explanation can be found in his work \cite{Magnusson2009}. The alignment is considered valid, if the mean likelihood over all input points is higher than a threshold  $P_{min}$.
\subsubsection{Pose Constraint}
\label{sec:poseconstraint}
The new pose estimate is calculated as: 
\begin{center}
$T^t_{MB_{estimate}} = \hat{T}^t_{MB} * T_{align}$
\end{center}
If the sanity check is successful, the pose constraint is added to the pose graph using a fixed covariance. For our experiments, the covariance was determined empirically.
\subsection{Curbstone Parameterization}
\label{sec:methodology:subsec:mapping}
Since curbstone maps can scale quickly given the multitude of possibly redundant observations, a memory overhead is induced. To reduce this memory footprint, we perform a curbstone parameterization. Since the map contains several artifacts like intersections or roundabouts, a curve parameterization was preferred over a polyline. Curbstones are not continuous throughout the whole map, as they often end at intersections. Thus, it naturally makes sense to split the map into single connected regions. In a first step, the raw curbstone pointcloud is subsampled. A clustering is then performed on the subsampled points, to find connected segments of a maximum length. The length-to-width ratio of each segment is then calculated. If a certain threshold is fulfilled, a Cubic B-Spline is fitted to the segment. By doing so, only the control points of the spline have to be saved, instead of all raw curbstone points. If the threshold is not fulfilled, the raw points are saved. The steps are explained in detail in the following and shown in Figure \ref{fig:schema_curbstone_param}.
\subsubsection{Subsampling}
\label{sec:methodology:subsec:subsampling}
The high point density of the raw pointcloud can result in high runtimes of the clustering as well as in overfitting of the spline to noise in the points. Thus, a spatial subsampling using a voxel grid with a leaf size of $30cm$ is performed. A pointcloud of the means of the points inside each voxel is then used for clustering.
\subsubsection{Clustering}
\label{sec:methodology:subsec:clustering}
The clustering is performed in a two-step fashion. First, a Euclidean clustering using a tolerance of more than $2m$ is performed to find large segments. Since the curvature can vary along long segments, fitting a single spline to it can be problematic, as different levels of detail are needed along the segment. An example is a curb going around a corner: While low curvature is desired in the straight sections, high curvature is needed in the area of the corner to properly describe the curb. Thus, the coarse cluster is split into smaller sub-clusters with a maximum expansion of 20$m$ before validating each sub-cluster by using an SVD Decomposition.
\subsubsection{Cubic B-Spline Fitting}
\label{sec:methodology:subsec:fitting}
Spline fitting usually refers to fitting a spline that goes through each single input point. However, due to the noisy nature of the curbstone segment, a best fit given a fixed amount of control points is preferred in this case instead of fitting every single point. To achieve this, the approach proposed by Wang et al. \cite{splinefitting} to fit an open cubic B-Spline is used. The number of control points is calculated proportionally to the approximate length of the segment, using $0.25 \ points/m$, but a minimum of $4$. For segments with a large width (indicating an intersection, road curve or round-about) a fixed amount of $20$ points is used to allow for a proper representation.
To validate our fitted spline, we define our goodness score $GS$ as follows: $$GS = \frac{\#Spline \ Inliers}{\#Spline \ Points} * \frac{\#Point \ Inliers}{\#Points}$$whereas $Spline Inliers$ is the number of sampled spline point close to a raw point and $Point Inliers$ the number of raw points close to points sampled from a spline.
Naively using all sub-segment points for the spline fitting can lead to an overfitting of the curve. Thus, the best set of points is found in a RANSAC-like manner. In each iteration, one third of the sub-segment points is sampled randomly. The spline is then fitted to the sampled points. Eventually, the spline with the highest score is chosen.
\subsubsection{Spline Sampling}
\label{sec:methodology:subsec:mapping}
To be able to perform matching with the input cloud, points are spatially uniformly sampled from the splines during runtime. Those sampled points are then used as the map pointcloud.
\section{Evaluation}
\label{sec:evaluation}
In the following section, the performance of the proposed pipeline is evaluated and compared against the \visloc pipeline as a benchmark.
Long-term experiments in an urban scenario are performed on varying weather and appearance conditions. A special focus is set on how curbstone map tracking influences localization accuracy and recall. Example cases are presented, where localization gaps in the \visloc pipeline could be bridged using curbstone localization. The sensor set-up of the \updrive vehicle and the datasets used in the experiments are described in the next section.

\subsection{The \updrive Platform}
\label{sec:eval:subsec:implemenation}
For the collection of the datasets, the \updrive vehicle was used. Its sensor setup consists of four fish-eye cameras, resulting in a surround view of the car. Gray-scale images with a resolution of $640 x 400 px$ are recorded at $30Hz$. Five Velodyne LiDARs are mounted on top of the car. Curbstones are obtained from the approach as described by Goga et al. \cite{Goga2018}. Additionally, a low-cost IMU and wheel tick encoders are used to provide odometry measurements. A consumer-grade GPS sensor is used to gain an initial position estimate and near-by map poses are used to generate an initial orientation estimate.
%
\subsection{\updrive Dataset Collection}
\label{sec:evaluation:subsec:datasets}
The \updrive dataset collection was recorded between December 2017 and November 2019 in Wolfsburg, Germany, at the Volkswagen factory and its surrounding area and aggregate a total driving distance of multiple $100$ kilometers. The environment is urban, with common artifacts such as busy streets, buses, zebra crossings and pedestrians. Since the data was collected over several months, seasonal appearance changes as well as multiple weather and day-time conditions are present. For this work, our dataset selection is dependent on the availability of curbstone measurements, which result from the curbstone detection pipeline from Goga et al. \cite{Goga2018}. Since the curbstone detection module was only enabled in some of our datasets, our evaluation dataset collection consists of $5$ sessions which totals to $10$ drives from August 2019 to November 2019. Each session contains two partially overlapping routes in opposite directions and consists of the same amount of sunny and cloudy/rainy conditions captured throughout a day. Recordings in rainy conditions are categorized as \textit{Cloudy}, since there is little difference in performance on rainy datasets as opposed to in dry conditions.
\subsection{Metrics}
\subsubsection{Localization Recall}
The fraction of the total travelled distance in which a successful localization was achieved is calculated as the localization recall $r$[\%].
While using only the visual pipeline, a localization attempt at time $t$ is accepted as successful if a minimum of $10$ inlier landmark observations is present after pose optimization. When using the combined pipeline, a localization is counted as successful, if the condition above is fulfilled or if a viable curbstone alignment (see section \ref{sec:methodology:subsec:map_tracking}) could be performed.
\subsubsection{Localization Accuracy}
As no ground-truth for the described dataset exists, the poses estimated by an \textit{RTK GPS} sensor are used instead as a reference. \textit{RTK GPS} altitude estimates are not reliable, thus the error in $z$ can not be calculated reliably. Therefore, we focus on the planar $\planarerror$ and lateral translation error $\laterror$ as well as on the orientation error $\orienterror$.

%
%
%
%

\newcommand{\ra}[1]{\renewcommand{\arraystretch}{#1}}
\begin{table*}\centering
\ra{1.3}
\scalebox{0.62}{
\begin{tabular}{@{}rrrrrcrrrrrcrrrrr@{}}\toprule
& \multicolumn{4}{c}{$\recallmt$\footnotesize{[\%]}} & \phantom{abc}& \multicolumn{4}{c}{ $\planarmedianerror$, $\latmedianerror$} &
\phantom{abc} & \multicolumn{4}{c}{$\orientmedianerror$}\\
\cmidrule{2-5} \cmidrule{7-10} \cmidrule{12-15}
& $10$-$08$ & $10$-$25$ & $11$-$08$  & $11$-$20$ && $10$-$08$ & $10$-$25$ & $11$-$08$  & $11$-$20$  && $10$-$08$ & $10$-$25$ & $11$-$08$  & $11$-$20$  \\ \midrule
$MOZARD$ - $Map$\\
$($08-21$)$ & 100.0 & 99.94 & 99.06 & 99.82 && 0.08 [0.34],0.04 [0.21]& 0.07 [0.26], 0.03 [0.13]& 0.09 [0.37], 0.04 [0.2]& 0.13 [0.37], 0.05  [0.2]&& 0.64 [0.75] & 0.74 [0.76] & 1.04 [1.33] & 1.09 [1.4]\\
$($08-21; 10-08$)$ & - & 100.0 & 100.0 & 100.0 && - & 0.06 [0.15], 0.03 [0.08] & 0.07 [0.24], 0.03 [0.13] & 0.1 [0.29], 0.04 [0.14] && - & 0.66 [0.65] & 1.15 [1.4] & 1.2 [1.4]\\
$($08-21; 10-08; 10-25$)$ & - & - & 100.0 & 100.0 && - & - & 0.07 [0.22], 0.03[0.11] & 0.09 [0.27], 0.03 [0.12] && - & - & 1.21 [1.43] & 1.23 [1.42]\\
$($08-21; 10-08; 10-25; 11-08$)$   & - & - & - & 100.0 && - & - & - & 0.07 [0.18], 0.02 [0.1] && - & - & - & 1.13 [1.38]\\
$VIZARD$ - $Map$\\
$($08-21$)$ & 100.0 & 98.2 & 97.94 & 91.76 && 0.08 [0.28], 0.04 [0.16] & 0.07 [0.26], 0.03 [0.13] & 0.09 [0.29], 0.04 [0.17] & 0.13 [0.37], 0.05 [0.21] && 0.64 [0.76] & 0.74 [0.76] & 1.1 [0.16] & 1.07 [1.28]\\
$($08-21; 10-08$)$ & - & 100.0 & 99.9 & 97.89 && - & 0.06 [0.13], 0.02 [0.07] & 0.07 [0.24], 0.03 [0.13] & 0.1 [0.29], 0.04 [0.14]&& - & 0.55 [0.67] & 1.15 [1.4] & 1.19 [1.37]\\
$($08-21; 10-08; 10-25$)$ & - & - & 100.0 & 99.18 && - & - & 0.07 [0.22], 0.03 [0.11] & 0.09 [0.25], 0.03 [0.13] && - & - & 1.21 [1.43] & 1.23 [1.41]\\
$($08-21; 10-08; 10-25; 11-08$)$  & - & - & - & 99.7 && - & - & - & 0.07 [0.18], 0.02 [0.1] && - & - & - & 1.13 [1.38]\\
\bottomrule
\end{tabular}
}
\caption{\label{table:loc_performance_summary}
The localization performance on the \updrive dataset, showing localization recall, and the median planar~$\planarmedianerror$, lateral~$\latmedianerror$ and orientation ($\orientmedianerror$) accuracy. The $90$~percentile is shown in square brackets. Numbering in round brackets defines the timestamp of the sessions used for mapping. E.g. (08-21) represents August, 21th.}
\end{table*}

\subsection{Localization Accuracy and Recall}
\label{sec:evaluation:subsec:localication_accuracy_and_recall}
%
%
%
In order to fully rely on \villoc to control the car in the UP-Drive project, a high localization recall with an accuracy below 0.5$m$ is paramount, as only short driving segments with no localization may be bridged with wheel-odometry before the car may deviate from its
designated lane.
Curbstones are not available for the whole trajectory, but for around 89\% of the distance of the map.
We compare localization recall and accuracy of our localization system to our prior work on visual localization - \visloc \cite{brki2019vizard}.
Note, however, that our prior work relied on the use of cameras for localization, as in contrast to the former, the latter is now able to use LiDAR and vision.
To demonstrate that curbstones provide useful additional information, we construct and expand a map iteratively using multiple datasets. Our first map is constructed from two datasets (one session) from August 2019. We then evaluate this map against multiple sessions from different months and add these sessions to our (multi-session) map in a iterative fashion. 
We present the resulting key evaluation metrics (localization recall $\recallmt$\footnotesize{[\%]}\normalsize\space
and localization accuracy) in Table~\ref{table:loc_performance_summary} over all sessions.
%
%
By including this comparison, we aim at highlighting the gain in localization recall attainable by using \villoc while keeping a consistent median translation and orientation error.
As shown in Table~\ref{table:loc_performance_summary}, \villoc is able to attain close to a $100\%$ recall performance on all $4$ sessions on the \updrive dataset, while \visloc performance increase correlates with the addition of sessions to its base map due to the change in visual appearance.
%
We further note that in both cases the planar median localization accuracy are below 15$cm$, while the median lateral error is below 10$cm$. The median orientation errors are on average less than 1 $degree$. For \villoc the 90th percentile shows an increase which is likely to be due to the higher uncertainty in precision of curbstone measurements. 
\subsection{Runtime}
\label{sec:eval:subsec:runtime}
On our live car platform Goga et al. \cite{Goga2018} demonstrated that their curbstone detection pipeline deployed on 2 Nvidia GTX 1080 takes around 20$ms$ for the CNN image segmentation to complete on all $4$ cameras. An additional 32$ms$ are needed for the fusion of 5 LiDARs to run on an Intel i7-3770K CPU. Our curbstone alignment module takes an average of approximately 25$ms$, while the map tracking module (with vision) can take from 27$ms$ with a single session map up to 48$ms$ on our largest multi-session map (see Table~\ref{table:loc_performance_summary}) and has been evaluated on an Intel Xeon E3-1505M CPU. This would allow \villoc to run with around 10$Hz$ on a single machine on a single session map. Table~\ref{table:loc_runtime_summary} summarizes our findings.
\begin{table}[h]
\begin{center}
\begin{tabular}{|c|c|} 
\hline
Module &  Average Runtime [ms]\\
\hline
Curbstone Detection & 52\\ 
Curbstone Tracking & 25\\ 
Map Tracking (\vislocns) & 27-48\\ 
\cline{1-2}
Total & 104-125\\ 
\hline
\end{tabular}
\end{center}
\caption{\label{table:loc_runtime_summary}
Runtime of each component of \villoc.
Curbstone Detection and Curbstone Tracking with average runtime over all evaluated datasets on a single session map, while Map Tracking shows average runtime for running on a single session map and on the largest multi-session map.
}
\end{table}
\subsection{Case Study}
\label{sec:eval:subsec:case_study}
\begin{figure}
\includegraphics[width=0.49\textwidth]{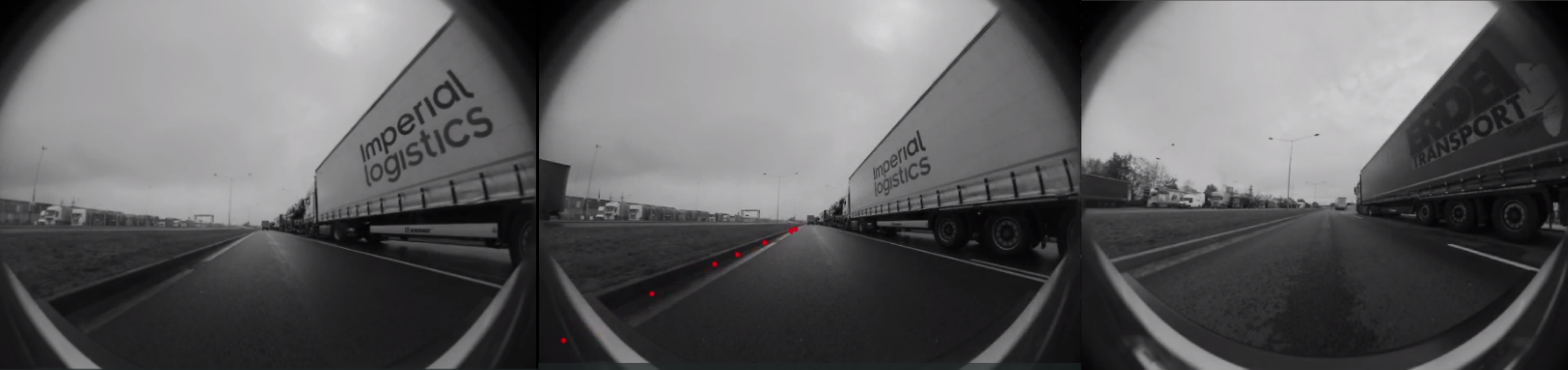}
\caption{\label{fig:fail_vis_vililoc_sample_images}
On the left, a sample image of a trajectory segment that fails to localize due to occlusion.
A lack of keypoints renders it unfeasible to match a sufficient number of map landmarks.
On the middle, the projected curbstone information is depicted in the camera frame in red - enabling a continued localization although visual localization failed.
On the right, a sample image is depicted where our curbstone and vision pipeline fail. In this case curbstones are actually detected but alignment fails due to our constraints.}
\vspace{-4mm}
\end{figure}

We provide further insights into our pipeline by showing specific failure examples for each component.
Sample images of a section where the visual localization fails on the evaluated datasets are depicted in Figure~\ref{fig:fail_vis_vililoc_sample_images}.
Due to occlusion and the absence of surrounding building structures, barely any stable visual cues are found in this section, preventing the visual localization system from matching a sufficient amount of landmarks from the map.
This example demonstrates the current limitations of \vislocns, while our \villoc pipeline is able to handle these sparse keypoint-based scenarios.
Unfortunately there are also scenarios where a lack of keypoints and curbstones exists or our curbstone alignment fails - hence conditions where both pipeline are likely to fail as depicted in the right image of Figure~\ref{fig:fail_vis_vililoc_sample_images}.
A further extension of our current framework to other geometric shapes such as poles, road markings could provide additional useful information that would allow us to further increase our localization performance.
Note that we used a single session map for the evaluation of this case study and \visloc is able to bridge some of these scenarios if enough datasets are provided during the mapping process.

\section{conclusions}
\label{sec:conclusions}
We presented \villocns, a geometric extension to our visual localization system for urban outdoor environments.
Through our evaluation on $8$ datasets, including several kilometers of real-world driving conditions, we demonstrated the benefits of using curbstone information for localization and mapping.
Our datasets used in the experiments contain challenging appearance conditions such as seasonal changes, wet road surfaces and sun reflections.
A comparison with our prior work demonstrated that we can achieve a higher recall performance while using less datasets during the mapping process, as the pipeline would fail due to sparse keypoint scenarios.
Our run-time analysis shows that our approach demonstrates real time capabilities. Although the curbstone detection stack of \villoc takes in average more computing time than \vislocns, it is to note that an object segmentation/detection algorithm on a self-driving car has to be deployed for environmental perception independent of whether a localization takes place or not. Even taking in account the total computational time, our approach still runs at $10Hz$ while needing up to four times less data while achieving the same localization performance.
We also showed specific cases where both of our pipelines would fail due to occlusions and/or curbstone misalignment giving suggestions for future work such as the extension of our approach to poles and road markings.
Our findings showed that by extending a keypoint based visual localization approach with geometric features - curbstones in our case, an improvement in robustness with consistent high accuracy in localization is obtained.

\addtolength{\textheight}{-10cm}   





\section*{ACKNOWLEDGMENT}

This project has received funding from the EU H2020 research project under grant agreement No 688652 and from the Swiss State Secretariat for Education, Research and Innovation (SERI) under contract number 15.0284.



{\small
\bibliographystyle{IEEEtran}
\bibliography{mendeley_2}
}

\end{document}